# Leveraging Contextual Relatedness to Identify Suicide Documentation in Clinical Notes through Zero Shot Learning

T. Elizabeth Workman, Ph.D.[1,2], Joseph L. Goulet, Ph.D.[3,6], Cynthia A. Brandt, M.D.[3,6], Allison R. Warren, Ph.D.[4], Jacob Eleazer, Ph.D.[4], Melissa Skanderson, M.S.W.[5], Luke Lindemann, Ph.D.[6], John R. Blosnich, Ph.D.[7], John O'Leary, M.Ed.[6,8], Qing Zeng-Treitler, Ph.D.[1,2]

*Abstract—* <u>Objectives</u>: Identifying suicidality including suicidal ideation, attempts, and risk factors in electronic health record data in clinical notes is difficult. A major difficulty is the lack of training samples given the small number of true positive instances among the increasingly large number of patients being screened. This paper describes a novel methodology that identifies suicidality in clinical notes by addressing this data sparsity issue through zero-shot learning. <u>Materials and Methods</u>: U.S. Veterans Affairs clinical notes served as data. The training dataset label was determined using diagnostic codes of suicide attempt and self-harm. A base string associated with the target label of suicidality was used to provide auxiliary information by narrowing the positive training cases to those containing the base string. A deep neural network was trained by mapping the training documents' contents to a semantic space. For comparison, we trained another deep neural network using the identical training dataset labels and bag-of-words features. <u>Results</u>: The zero shot learning model outperformed the baseline model in terms of AUC, sensitivity, specificity, and positive predictive value at multiple probability thresholds. In applying a 0.90 probability threshold, the methodology identified notes not associated with a relevant ICD-10-CM code that documented suicidality, with 94% accuracy. <u>Conclusion</u>: This new method can effectively identify suicidality without requiring manual annotation.

## I. Introduction

Suicide is a significant problem in the United States, increasing 35.2% from 1999 to 2018, and from 10.5 to 14.2 suicides per every 100,000 individuals in that same time period [1] In 2020, 45,979 people died from suicide, and approximately 1.2 million attempted suicide in the United States [2] Its estimated cost is over $70 billion annually in lost productivity and medical care [3]; this calculation does not include residual costs from the estimated 4-17 people closely tied to the suicide decedent who are left bereaved [4]. Suicide, however, is a complicated problem that includes a dynamic web of individual-level risk factors (e.g., depression, substance use behaviors, personality traits), interpersonal risk factors (e.g., violence, victimization), and community-level factors (e.g., unemployment, stigmatization of mental illness) [5, 6].

Veterans are especially affected by suicide, with an age- and sex-adjusted rate that is 1.5 times higher than nonveterans [7]. The Department of Veterans Affairs (VA) operates the single largest integrated health care system in the U.S., and has devoted resources to suicide prevention, including the Suicide Prevention Applications Network (SPAN), embedding suicide prevention coordinators and special reporting measures in facilities [8], increased mental health staffing, partnerships with community care organizations, and enhanced surveillance and monitoring through its electronic health record (EHR) system [9, 10]. Additionally, the VA has continual efforts to develop predictive analytics to identify patients at the highest risk of suicide [8, 11] The data elements for these predictive analytic algorithms rely on structured data (e.g., International Classification of Disease [ICD] diagnosis codes, prescription data, socio-demographic data, care utilization metrics) [12] which often provide an incomplete record [13, 14]. Less is known about how unstructured data, such as contained in clinical notes, can contribute to suicidality (i.e., suicidal ideation or attempt) identification and prevention. Given that a suicide attempt is one of the greatest risk factors for subsequent suicide death, a more thorough means of detecting such events is warranted [15].

### A. Background and Significance

Natural language processing (NLP) combined with machine learning may add value to suicide documentation research. Supervised machine learning methods use "supervised", or pre-classified data. However, naïve attempts at note retrieval using keyword search alone quickly demonstrate the difficulty of this problem, as words such as "suicide" occur in standard questionnaires which are included in many notes, with few actually documenting suicidality. For instance, in a prior experiment we carried out, we randomly collected 1,000 VA notes containing the term "suicidal" or "suicide" from 1,000 individual patients and performed manual chart review for affirmed suicidality. Only 1.57% of these notes documented actual suicidality. Patient reluctance to disclose suicidal ideation provides a further complicating factor [16, 17]. As a result, a patient's negative response to a suicide ideation inquiry may not reflect their real feelings or intentions. Additionally, relying on structured data alone will result in incomplete identification of patients who have or are experiencing suicidality, because relevant coding is prone to underuse [8]. However, not all clinical notes associated with relevant structured data document suicidality. For example, a note documenting a secondary service such as group therapy, or a note documenting fluid intake may not directly document suicidality.

[1]Biomedical Informatics Center; The George Washington University; Washington DC, USA; [2]VA Medical Center, Washington, DC, USA; [3]Department of Emergency Medicine, Yale School of Medicine, New Haven, CT, USA; [4]PRIME Center, VA Connecticut Healthcare System, West Haven, CT, USA; [5]Research, VA Connecticut Healthcare System, West Haven, CT, USA; [6]VA Connecticut Healthcare System, West Haven, CT, USA; [7]Suzanne Dworak-Peck School of Social Work, University of Southern California, Los Angeles, CA, USA; [8]Department of Internal Medicine, Yale School of Medicine, West Haven, CT;

Prior attempts to apply NLP and machine learning are often limited to mental health-oriented notes and may suffer if using imbalanced data. Levis et al.[18] applied sentiment analysis and various machine learning algorithms to classify suicide, using VA psychotherapy notes, yielding area under the curve (AUC) ratings comparable to chance. Fernandes et al.[19] obtained excellent NLP performance in their study of clinical notes from the Clinical Record Initiative Search (CRIS), but performance was computed after removing neutral (non-suicide) results from their machine learning output. Carson et al. enriched notes associated with suicide attempt that were then used to train a random forest model achieving 83% sensitivity, but only 22% specificity [20]. Cook et al. [21] applied a bag-of-words approach with machine learning to identify suicide ideation and psychiatric symptoms using notes for patients identified as having performed self-harm, achieving 61% PPV (positive predictive value), 59% sensitivity, and 60% specificity, with results varying depending on the task. Zhang et al. sought to identify psychological stressors using a pre-annotated dataset of psychiatric evaluation records from the CEGS N-GRID 2016 challenge [22] as a gold standard, for a conditional random fields machine learning model, [23] yielding final F scores of 73.91% and 89.01%, respectively, on exact and inexact stressor matching, and 97.73% and 100% respectively, for exact and inexact suicide recognition on instances of the positive keywords with the stressors; however, their evaluation methods for this are not detailed.

Zhong et al. applied structured data and NLP to identify suicidal behavior in pregnant women, achieving PPV of 76% and 30%, for women identified through relevant diagnostic codes and through NLP for women not receiving a relevant diagnostic code, respectively [24]. Obeid et al.[25] trained a convolutional neural network that achieved an AUC of 0.882 and an F1 score of 0.769 in predicting relevant suicide ICD codes in subsequent years. Using notes from psychiatric encounters, Cusick et al. [26] developed a rule-based NLP tool to identify positive instances of suicide-oriented keywords that leveraged NegEx. [27] They also developed different weakly-supervised machine learning models. A convolutional neural network receiving Word2Vec [28] word embeddings as input achieved precision, recall, F1 score, and AUC values of 0.81, 0.83, 0.82, and 0.946. In a subsequent evaluation the convolutional neural network correctly classified 87% of the 23 notes (of 5000 clinical notes) receiving a positive classification, from notes for patients diagnosed with depression or prescribed an antidepressant. In a related task Tsui et al. [29] used prior structured and unstructured data (clinical notes from history, physical examination, progress notes and discharge summaries) of inpatient and emergency room patients with a coded suicide attempt, to identify first-time suicide attempts in a case-control study. An ensemble of extreme gradient boosting (EXGB) yielded best performance, with an AUC ranging from 91.9% to 93.2%, according to time window between prior data and suicide attempt diagnosis. Recently, Rozova et al. obtained promising results (87% AUC) using a gradient boosting model, although the study was limited to emergency room triage notes [30].

Seeking suicidality *in all types of clinical notes*, among *all types of patients*, or when hampered by *imbalanced data*, is indeed a complex task. Some of the methods in the papers cited above tend to suffer from low precision, specificity, and possibly also low sensitivity (recall). Identifying probability thresholds addresses these problems, providing flexibility for a given task. For example, a high probability threshold (e.g., the top ten percent) can serve as a means for identifying documentation indicating suicidality and its risk with high precision. When the prevalence is very low, which is often the case of true positive suicidality documentation, the optimal threshold needs to balance metrics such as the true positive rate (sensitivity, also known as recall), specificity, and the positive predictive value (precision). A strategic implementation of a technique like Zero-Shot Learning may also provide accurate identification of suicidality in clinical notes.

### B. Zero-Shot Learning

Zero-Shot Learning (ZSL) enables predictions on unseen data using a model trained on data that has labels that are different than those of the unseen data [31, 32]. It largely operates by mapping select properties of the data (i.e., the "feature space") to a semantic representation (i.e., the "semantic space") that enables prediction of unseen classes [33]. In other words, auxiliary information must be provided on the labels of the unseen classes to make it possible for a trained model to recognize them in the testing data.

ZSL has been applied in several computer vision tasks [34, 35], as well as NLP tasks [36]. Accordingly, a feature space can consist of data derived from images [37] or text [36]. The semantic representation can be based on several different approaches, including data attributes, semantic word vectors as those provided by skip-gram or continuous-bag-of-word architectures [33] or BERT output [38], or knowledge graphs [33]. Examples in NLP applications include semantic utterance classification [39] multilingual translation [40] and emotion detection [41]. However, other than Sivarajkumar and Wang's work [38] there is little ZSL research in unstructured clinical text data.

Naturally, different semantic representations affect the accuracy of ZSL [42]. In this study, we leveraged word embedding and usage context.

### C. Objectives

We investigated a ZSL methodology applied to a binary suicidality classification task. The training dataset was constructed using diagnostic codes (ICD-10-CM codes) related to suicide. Our target label is the broader concept of suicidality. To enable ZSL, a base string representing suicidality was selected. We then built the semantic space by identifying key features associated with suicidality in the training dataset. A DNN model was developed using the training data and tested on two different sets of unseen data with the unseen label of suicidality. Specifically, we sought to answer:

- Will ZSL effectively identify suicidality documentation from among all types of clinical notes, using review by clinicians as the reference standard?
- Will ZSL effectively identify suicidality or suicide risk documentation from among clinical notes not associated with a relevant ICD-10-CM code, by probability threshold, in terms of precision, using the same reference standard?

We are unaware of previous descriptions of this methodology and to our knowledge it has not been used prior to this study.

## II. METHODS

### A. Training Data

A training dataset was created using two corpora. The first corpus consisted of 50,000 randomly selected VA clinical notes from outpatient encounters recorded between 2016 and 2019 which contained the base string "suicid" (e.g. "suicide", "suicidal" ) and were associated with at least one ICD-CM-10 code identified by the National Health Statistics Report from the Centers for Disease Control and Prevention (CDC) indicating suicide attempt or intentional self-harm.[43] This corpus is referred to as *stringAndDx* (9170 unique patients). The second corpus consisted of 50,000 randomly selected VA clinical notes from outpatient encounters recorded between 2016 and 2019 that were associated with other ICD-CM-10 codes that were irrelevant to suicidality or self-harm. These notes were extracted from patients matching the *stringAndDx* patients in age (at the time of document retrieval), race, and ethnicity. This second corpus is referred to as *noDx* (8638 unique patients). Each corpus was preprocessed by transforming all letters to lower case, removing basic formatting markup and punctuation, separating character strings into tokens (words), separating relevant concatenated tokens (e.g., "suicidalhomicidal" to "suicidal" "homicidal"), and removing all tokens that did not entirely consist of letters.

### B. Semantic Space Feature Extraction and Mapping

The task to build the semantic space was carried out in three steps: First, we identified a list of features that are potentially relevant for the positive training label. Second, we created word embeddings using a skip-gram architecture. Third, we identified context words of the selected features using the word embeddings. In a fourth step, a contextual weight is assigned to each feature for each document in mapping the semantic space to the feature space.

In the first step, inverse document frequency (TF-IDF) analysis was used to identify the *n* most important terms in each corpus. For this investigation, *n* = 1000. TF-IDF evaluates term frequency using the count of documents containing a given term. In each document, the relative frequency of each term is weighted by the log of the number of documents in the corpus divided by the number of documents containing the term, as shown in (1)

$$t_{i,j} = tf_{i,j} * log(\frac{n}{df_i}) \quad (1)$$

where $t_{i,j}$ is term i in document j, $tf_{i,j}$ is the relative frequency of term i in document j, n is the total number of documents, and $df_i$ is the number of documents containing term i. Because TF-IDF is a document-based measurement, we used the mean TF-IDF value for each term in its respective corpus. The words with the top TF-IDF scores that are unique to the *stringAndDx* corpus were treated as features. Figure 1 illustrates this process. Each circle represents terms from one of the corpora. Sets *a* and *b* are the words with the top *n* TFIDF scores for *stringAndDx* and *noDx*, respectively. Set *c* is the overlap between *a* and *b*. The feature set F contains words that are in set *a*, but not in the overlap set *c* or in set *b* (f ∈ *a* and f ∉ *c* and f ∉ *b*).

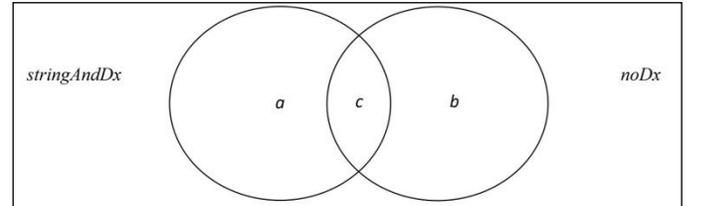

Figure 1. Feature identification. Words that are deemed as features are in set *a*, excluding words in *c* and *b*.

In the second step, we created a Word2Vec model using the *stringAndDx* corpus. In this study, the model was a shallow neural network with the hidden layer containing 300 nodes, applying the skip-gram architecture, with an analytic window size of 5, trained through 10 iterations.

In the third step, we identified the top *m* context words for each feature word using the word embeddings from the Word2Vec model. The *m* words most similar to each feature word, according to cosine similarity values, served as its context words. In this investigation, *m* = 50.

In the fourth step, we map the feature space, i.e. a document's preprocessed content, to the semantic space. A weight *v* is assigned to each feature word for each document, based on its occurrence with its context words in a window in the document's text. This weight is the summed total of the cosine similarity between the feature and a co-occurring context word multiplied by the mean TF-IDF value of the feature word. The formula is shown in (2)

$$v = \sum_{x \in F, y \in D} cosSim(x,y) * tfidf(x) \quad (2)$$

where x is a feature in F, the set of features in the semantic space, and y is a context word of set D, the context words for x in the semantic space, which occurs in a five-word window around x in the document's text. This process is illustrated in Figure 2, where "pattern" (highlighted in light gray) is a feature word, and "internalizing" and "fitful" (highlighted in dark gray) are among its set of context words and appear in a five-word window.

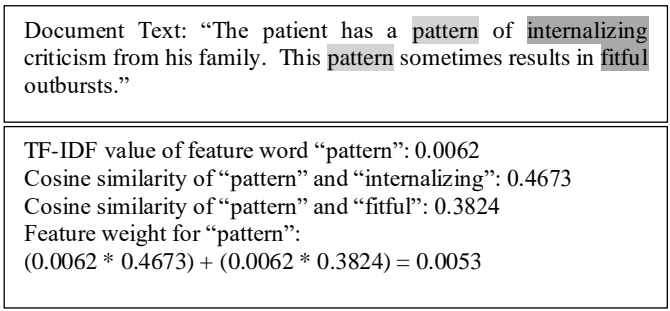

Document Text: "The patient has a pattern of internalizing criticism from his family. This pattern sometimes results in fitful outbursts."

TF-IDF value of feature word "pattern": 0.0062
Cosine similarity of "pattern" and "internalizing": 0.4673
Cosine similarity of "pattern" and "fitful": 0.3824
Feature weight for "pattern":
(0.0062 * 0.4673) + (0.0062 * 0.3824) = 0.0053

Figure 2. Example of deriving a feature weight using (2)

If a feature word is not in the text, its value is zero for the given document.

### C. Model Development

20,000 documents were randomly selected from each corpus (*stringAndDx* and *noDx*). We trained a DNN model (here referred to as the ZSL DNN) consisting of five fully-connected hidden layers of alternating sizes of 30 or 70 nodes, with each layer implementing a dropout rate of 0.5. We implemented the Adam optimizer [44], with a learning rate of 0.0012, beta 1 value of 0.92, beta 2 value of 0.9992, and an epsilon value of 1e-08, with binary cross entropy as the loss function, and the sigmoid function in the output layer, since it was a binary classification task. The architecture and hyperparameters were chosen on empirical grounds, after experimentation. Each document from the *stringAndDx* corpus was classified as "1" (a generic positive instance), and each document from the *noDx* corpus was classified as "0" (a generic negative instance). These labels do not indicate whether or not the given document directly pertains, or not pertains, to suicidality or its risks, but an association with a structured data element, and for those labeled "1", also containing a base string. Balancing the positive and negative approximated training datasets in this manner (i.e., providing balanced training examples) addressed the problematic issue of otherwise training a model with few positive and many negative instances. We implemented a 60% training, 20% validation, and 20% testing split in developing the ZSL DNN. Figure 3 illustrates the method.

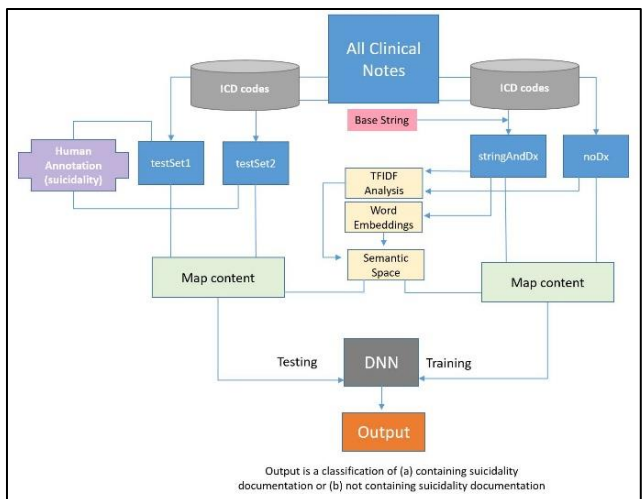

Figure 3. Method. The corpora *stringAndDx* (2016-19), *noDx* (2016-19), *testSet1* (2020), and *testSet2* (2020) are unique and extracted from all clinical notes based on associated ICD-10-CM codes, and in the case of *stringAndDx*, where a base string is also present; corpora content is preprocessed. The *stringAndDx* and *noDx* corpora are used in the TF-IDF analysis to identify feature words that are unique to *stringAndDx* (step 1). *stringAndDx* is applied to a skip-gram model to produce word embeddings (step 2). Feature words and their significant context words (determined through the word embeddings) form the semantic space (step 3). The contents of *stringAndDx* and *noDx* are mapped to the semantic space, using a function to determine feature word weights (step 4). The mapped contents of *stringAndDx* and *noDx* documents are used to train the ZSL DNN, using generic labels 1 and 0, respectively. The mapped contents of unseen *testSet1* and *testSet2* notes were classified by the trained ZSL DNN, for the classes (a) containing suicidality documentation, or (b) not containing suicidality documentation. Human annotation independently classified random documents from *testSet1* and *testSet2* for the same classes (a) containing suicidality documentation, or (b) not containing suicidality documentation; human annotation also assessed documents from *testSet2* containing the base string that received a probability of 0.90 or greater, for these classes and suicidality risk factors.

### D. Evaluation

The authors randomly retrieved 5,000 different clinical notes recorded in 2020 that were associated with at least one of the relevant IDC-10-CM codes. This corpus is subsequently labeled as *testSet1*. The authors also randomly retrieved 5,000 different clinical notes recorded in 2020 that were associated with other IDC-10-CM codes irrelevant to suicidality or self-harm. This corpus is subsequently labeled *testSet2*.

The contents of each of the notes in *testSet1* and *testSet2* were mapped to the semantic space, i.e., deriving a weight for each feature word as described earlier in the fourth step. Then, the trained ZSL DNN was used to classify the notes in *testSet1* and *testSet2* as (a) containing suicidality documentation, or (b) not containing suicidality documentation.

In joint sessions, two clinical psychologists familiar with VA clinical note documentation together identified suicidality (i.e., current or past suicide ideation or attempt) in 200 notes randomly selected from *testSet1* and *testSet2* (100 from each test set), after being instructed to look for documentation for these specific events. They addressed differences of opinion through discussion and mutual consensus during the joint sessions. In a second evaluation, to explore how the application's output may serve to identify patients who had experienced or were at risk for suicidality, but never formally diagnosed as such, the clinicians examined the *testSet2* notes containing the base string "suicid" that received a probability value of 0.90 or greater from the trained ZSL DNN, for documentation of suicidality and/or its risk factors, according to NIH guidelines.[45] This threshold was chosen in order to explore how high-probability documents (i.e. the top 10% in terms of probability) would be representative in identifying documented suicidality or its risk factors with high precision, thus addressing our second question.

*1) Baseline Comparison*

For comparative purposes, the 163 most frequent bigrams unique to the *stringAndDx* corpus were identified and used in a bag-of-words baseline model. We trained a DNN (here referred to as the Baseline DNN) using these 163 bigrams as features for the 20,000 *stringAndDx* documents and the 20,000 *noDx* documents. This baseline DNN was also used to classify the notes in *testSet1* and *testSet2*, for (a) containing suicidality documentation, or (b) not containing suicidality documentation, using the 163 most frequent bigrams as features.

## III. RESULTS

The first step of the new method (described in Methods) identified 163 feature words associated with suicidality diagnosis. The top thirty feature words are listed in Table I. No form of the base string "suicid" was found among the 163 final feature words. Both "suicide" and "suicidal" were prominent terms in both the *noDx* and *stringAndDx* corpora, along with terms like "psychiatrist" and "psychosocial"; this is likely due to the proliferation of objects like questionnaires, and mental health care documentation in notes that are unrelated to suicidality.

TABLE I
TOP 30 FEATURE WORDS

| flag | overdose | coordinator | took | spc |
|---|---|---|---|---|
| observation | called | warning | pills | prf |
| unknown | interrupted | gun | placement | lcsw |
| lethal | outcome | reportedly | notified | sdv |
| occurred | police | protocol | od | supports |
| seeking | category | preparatory | cut | determined |

### A. ZSL DNN and Baseline DNN Performance

The classifications by the clinicians and the probabilities assigned by the ZSL DNN and the Baseline DNN were first assessed by AUC score. The results are in Table II and Figure 4.

TABLE II
AUC PERFORMANCE

| ZSL DNN | Baseline DNN |
|---|---|
| 0.946 | 0.47 |

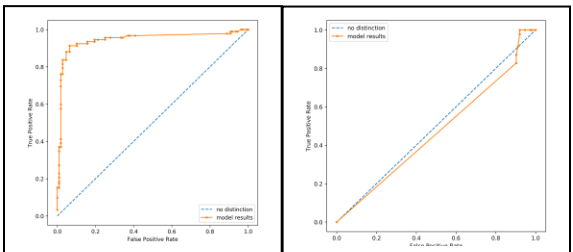

Figure 4. ZSL DNN AUC results (left), Baseline DNN AUC results (right)

In terms of AUC, the ZSL DNN trained through mapping the semantic space to the feature space outperformed the Baseline DNN trained with the bigram bag-of-words features.

The sensitivity, specificity, and PPV results at 0.15, 0.5, and 0.85 probability thresholds for each DNN are in Tables III-V. Probability refers to the probability the DNN assigned to each note for positive suicidality documentation. We applied the median probability (0.1499, rounded) assigned by the ZSL DNN to the *testSet2* documents (the test set containing random notes associated with irrelevant ICD-10-CM codes) in forming minimum and maximum thresholds; 0.5 is a standard midpoint probability threshold. The combined scores in these tables were computed with all true positives, true negatives, false positives, and false negatives for both test sets, for the indicated metrics. Values of NaN (not a number) occurred where there were no true positives or false positives.

TABLE III
EVALUATION RESULTS AT 0.15 PROBABILITY THRESHOLD

| ZSL DNN | Sensitivity/Recall | Specificity | Precision/PPV |
|---|---|---|---|
| *testSet1* | 97% | 100% | 91% |
| *testSet2* | 100% | 64% | 05% |
| Combined | 97% | 59% | 67% |
| **Baseline DNN** | | | |
| *testSet1* | 99% | 0% | 90% |
| *testSet2* | 50% | 09% | 01% |
| Combined | 98% | 08% | 48% |

TABLE IV
EVALUATION RESULTS AT 0.5 PROBABILITY THRESHOLD

| ZSL DNN | Sensitivity/Recall | Specificity | Precision/PPV |
|---|---|---|---|
| *testSet1* | 92% | 40% | 93% |
| *testSet2* | 50% | 97% | 25% |
| Combined | 91% | 92% | 90% |
| **Baseline DNN** | | | |
| *testSet1* | 92% | 0% | 89% |
| *testSet2* | 50% | 10% | 1% |
| Combined | 91% | 9% | 46% |

TABLE V
EVALUATION RESULTS AT 0.85 PROBABILITY THRESHOLD

| ZSL DNN | Sensitivity/Recall | Specificity | Precision/PPV |
|---|---|---|---|
| *testSet1* | 77% | 70% | 96% |
| *testSet2* | 50% | 100% | 100% |
| Combined | 76% | 97% | 96% |
| **Baseline DNN** | | | |
| *testSet1* | 0% | 100% | NaN/div by 0 |
| *testSet2* | 0% | 100% | NaN/div by 0 |
| Combined | 0% | 100% | NaN/div by 0 |

The ZSL DNN outperformed the Baseline DNN in most metrics at all probability thresholds.

### B. Second Evaluation

To explore how this new methodology can identify clinical notes documenting suicidality that are not associated with a relevant ICD-10-CM code with high precision, the clinicians also reviewed the 16 notes from *testSet2* containing the base string "suicid' that received a probability at or above 0.90 from the trained ZSL DNN. The clinicians noted suicide ideation or attempt, and the presence of the following suicide risk factors, based on National Institute of Mental Health guidelines [45]:

- Depression and other mental health disorders
- Substance abuse disorder
- Family history of a mental health or substance abuse disorder
- Family history of suicide
- Family violence, including physical or sexual abuse
- Having guns or other firearms in the home
- Being in prison or jail
- Being exposed to others' suicidal behavior

Of these 16 clinical notes (associated with 16 different patients), 7 documented current or past suicide ideation or attempt. Eight of the remaining notes included one or more risk factors for suicide (nearly all included multiple risk factors). In all, 15 of the 16 notes contained documentation of current or past suicide ideation or attempt, and/or suicide risk factors, for patients who had never received a suicidality ICD-10-CM code diagnosis during the study period, achieving a PPV of 93.8%.

## IV. Discussion

Regarding the study's original questions, our ZSL approach effectively identified suicidality in all types of clinical notes, surpassing the performance of the bag-of-words baseline in conjunction with deep learning. It also effectively identified suicidality or suicide risk documentation from among clinical notes not associated with a relevant ICD-10-CM code with high precision, on probability threshold.

### A. Semantic Space

In this work, the semantic space development is framed as feature extraction where mapping is enhanced by attaching weights to features found in the data, an approach also used in computer vision ZSL [46]. The semantic space captures natural data properties by identifying salient terms and relevant contextual terms in collective clinical suicidality documentation (i.e., a corpus of notes associated with relevant ICD codes). Table 1 lists 30 prominent feature words associated with collective suicidality documentation after removing terms associated with other kinds of documents. There is an intuitive sense to these words; "flag" is found in the phrase "high risk for suicide flag"; "overdose" and "cut" refer to suicide methods; "pills" and "gun" refer to suicide instruments. Identifying terms contextually similar to these provides patterns in relevant documentation. Again, this has an intuitive logic. The most contextually similar terms to "flag" include "reactivate" and "deactivate" (for a high suicide risk flag) and "high" (the level of risk). The most contextually similar terms to "pills" include "handful", "fistfuls", and "bunch", implying large quantities, along with "overdosing" and "took", the associated actions. The feature word "spc" indicates VA's suicide prevention coordinators, which is a structural change that VA implemented for suicide prevention [10]. Concordantly, "police" and "lcsw" (i.e., licensed clinical social worker) refer to other professions highly associated with individuals at risk for suicide. For example, police may be activated for a rescue, and a licensed clinical social worker may be involved in treatment planning or referral connections for suicidal individuals. The feature words "prf" and "sdv" refer to "patient record flag" and "self-directed violence", respectively. The semantic space provided an efficient representation for effective mapping to the feature space.

### B. Data Retrieval and Model Training

Using associated structured data elements like ICD-10-CM codes, and a base string provides a means to locate equally sized corpora for training that could be generically labeled "0" or "1". These labels were primarily based on a structured data association, since their individual unstructured content was mostly unknown. This approach solves the issue of imbalanced training data. The predominant clinical note types (Appendix) also illustrate this. Most of the frequent note types associated with one of the relevant CDC ICD-10-CM codes and containing the base string are relevant to suicidality. Addendum is a common note type [47] associated with many domains [48]. The most frequent note types not associated with a relevant code resemble frequencies of all note types in the VA [47].

### C. Identifying Suicidality Documentation

To our knowledge, this method has not been applied in other studies. Unlike VA surveillance methods using structured data, it also leverages information found in EHR notes. Also, unlike other NLP methods [18, 20, 21, 23, 24, 26, 29, 30] it can be applied to all patients and note types. In other studies, a bag-of-words approach has been applied to suicidality identification and other machine learning tasks [21, 49, 50]. However, the results of this current study suggest that the complexity of suicidality documentation demands a more targeted approach.

This method could complement existing measures like SPAN, alerting suicide prevention coordinators of additional patients at risk. The results of the two clinical psychologists' evaluations demonstrate the method's efficiency in identifying suicidality documentation for documents where there is no relevant ICD-10-CM code. The performance on both test sets demonstrates the methodology's effectiveness in classifying notes that are mixed in terms of ICD-10-CM coding.

Tables III - V suggest that the probability threshold can be adjusted to suit a specific task like finding suicidality and its risk factors with high precision among notes not associated with a relevant ICD-10-CM code. This is especially true considering the small prevalence of suicidality documentation in clinical notes. The second evaluation (which yielded 93.8% PPV) demonstrates this. By applying a high probability threshold of 0.90 to all 5000 *testSet2* documents and focusing on clinical notes containing the base string, of the 16 documents (for 16 different patients), 94% contained suicidality and/or suicidality risk factor documentation, based on clinician review. These results exceed those of Cusick et al.'s [26] similar task, where 87% of notes were correctly classified, among notes for patients diagnosed with depression or prescribed an antidepressant. In this current study's second evaluation, none of the 16 patients identified had ever received a suicide ICD-10-CM code during the study's time period. It is impossible to know if the patients in the 8 notes simply containing documented risk factors were suicidal or not based solely on electronic health records. Suicidal patients sometimes deny suicide ideation or attempt [16, 51]. For example, in one note from the chart review associated with a relevant ICD-10-CM code, the patient reportedly denied suicide ideation, even after checking into the hospital hours earlier for a self-reported suicide attempt.

### D. Future Work

This work is part of a larger study of patients at risk for suicide.[52] The next step is to combine these findings with prior work. We also plan an analysis of patients from first suicide ideation or attempt documented in the VA system, to understand their evolution of care.

### E. Limitations

VHA data largely cover a population of older men. However, the amount of women and younger patients is increasing, thus also increasing the generalizability of these findings. The corpora retrieval method we used to train the ZSL DNN is dependent on clinicians' use of the relevant ICD-10-CM codes in documenting care, which may be prone to underuse [8]. However, the results of this study indicate the method's utility. Due to environmental computational limitations, we randomly selected 20,000 notes from the *stringAndDx* corpus, and 20,000 notes from the *noDx* corpus for training the ZSL DNN.

## V. Conclusion

We developed a new methodology to identify suicidality in clinical notes using zero-shot learning (ZSL). A trained ZSL deep neural network (DNN) outperformed a DNN trained using a baseline bag-of-words method in AUC scores and other metrics assessed at various probability thresholds on unseen data, according to expert review. This novel methodology identifies suicidality and its risk factors with high precision, when applying a 0.90 probability threshold, in VA clinical notes not associated with a relevant ICD-10-CM code. This methodology could complement existing suicidality identification measures. These findings hold promise for future research.

## Appendix

**Most Frequent Note Types in Training Data by Corpus**

| *stringAndDx* | | *noDx* | |
|---|---|---|---|
| Note Type | Count | Note Type | Count |
| Addendum | 2844 | Addendum | 5683 |
| Suicide Behavior and Report | 843 | Primary Care Secure Messaging | 291 |
| Suicide Prevention Telephone Note | 811 | Nursing Note | 228 |
| Suicide Behavior and Overdose Report | 613 | Administrative Note | 207 |
| Suicide Prevention Note | 452 | State Prescription Drug Monitoring Program | 110 |
| Suicide Prevention Safety Plan | 448 | Care Flow Sheet | 88 |
| Mental Health Nursing Assessment Note | 374 | Telephone Contact | 75 |
| Veterans Crisis Line Note | 222 | Mental Health Diagnostic Study Note | 71 |
| Social Work Note | 213 | Non VA Care Consult Result Note | 69 |
| Suicide Prevention Contact | 212 | Operation Report | 64 |

### Acknowledgment

The views expressed are those of the authors and do not necessarily reflect those of the Department of Veterans Affairs, the United States Government, or the academic affiliate institutions. This work was funded by Veterans Affairs Health Services Research and Development Services grant IIR 18-035 Understanding Suicide Risks among LGBT Veterans in VA Care, and NIH National Center for Advancing Translational Sciences grant UL1TR001876.